\documentclass{bmvc2k}

\title{Complex Scene Image Editing by Scene Graph Comprehension}

\addauthor{Zhongping Zhang}{zpzhang@bu}{1}
\addauthor{Huiwen He}{huiwenhe@bu.edu}{1}
\addauthor{Bryan A. Plummer}{bplum@bu.edu}{1}
\addauthor{Zhenyu Liao}{zyliao@amazon.com}{2}
\addauthor{Huayan Wang}{wanghy514@gmail.com}{3}

\addinstitution{
 Boston University\\
 MA, USA
}
\addinstitution{
 Amazon\\
 CA, USA
}
\addinstitution{
 Kuaishou Technology\\
 CA, USA
}
\makeatletter
\DeclareRobustCommand\onedot{\futurelet\@let@token\@onedot}
\def\@onedot{\ifx\@let@token.\else.\null\fi\xspace}

\def\eg{\emph{e.g}\onedot} \def\Eg{\emph{E.g}\onedot}

\def\etal{\emph{et al}\onedot}
\makeatother

\definecolor{lightmauve}{rgb}{0.86, 0.82, 1.0}
\definecolor{lightgoldenrodyellow}{rgb}{0.98, 0.98, 0.82}
\definecolor{lightapricot}{rgb}{0.99, 0.84, 0.69}
\definecolor{lightblue}{rgb}{0.55, 0.85, 0.9}
\definecolor{lightskyblue}{rgb}{0.53, 0.81, 0.98}
\definecolor{non-photoblue}{rgb}{0.64, 0.87, 0.93}
\definecolor{lightcornflowerblue}{rgb}{0.6, 0.81, 0.93}
\definecolor{lightgreen}{rgb}{0.56, 0.93, 0.56}
\definecolor{lightseagreen}{rgb}{0.13, 0.7, 0.67}
\definecolor{lightpink}{rgb}{1.0, 0.71, 0.76}

\def\eg{\emph{e.g}\bmvaOneDot}
\def\Eg{\emph{E.g}\bmvaOneDot}
\def\etal{\emph{et al}\bmvaOneDot}

\runninghead{Zhang \lowercase{\emph{et al.}}}{Complex Scene Image Editing by Scene Graph Comprehension}

\usepackage{graphicx}
\usepackage{amsmath}
\usepackage{amssymb}
\usepackage{booktabs}
\usepackage{multirow}
\usepackage{enumitem}

\begin{document}

\maketitle

\begin{abstract}
Conditional diffusion models have demonstrated impressive performance on various tasks like text-guided semantic image editing. Prior work requires image regions to be identified manually by human users or use an object detector that only perform well for object-centric manipulations. For example, if an input image contains multiple objects with the same semantic meaning (such as a group of birds), object detectors may struggle to recognize and localize the target object, let alone accurately manipulate it. To address these challenges, we propose a two-stage method for achieving complex scene image editing by Scene Graph Comprehension (SGC-Net). In the first stage, we train a Region of Interest (RoI) prediction network that uses scene graphs and predict the locations of the target objects. Unlike object detection methods based solely on object category, our method can accurately recognize the target object by comprehending the objects and their semantic relationships within a complex scene. The second stage uses a conditional diffusion model to edit the image based on our RoI predictions. We evaluate the effectiveness of our approach on the CLEVR and Visual Genome datasets.  We report an 8 point improvement in SSIM on CLEVR and our edited images were preferred by human users by 9-33\% over prior work on Visual Genome, validating the effectiveness of our proposed method. Code is available at \href{https://github.com/Zhongping-Zhang/SGC_Net}{github.com/Zhongping-Zhang/SGC\_Net}.
\end{abstract}

\section{Introduction}
\label{sec:intro}

\begin{figure}[t]
    \centering
    \includegraphics[width=.75\linewidth]{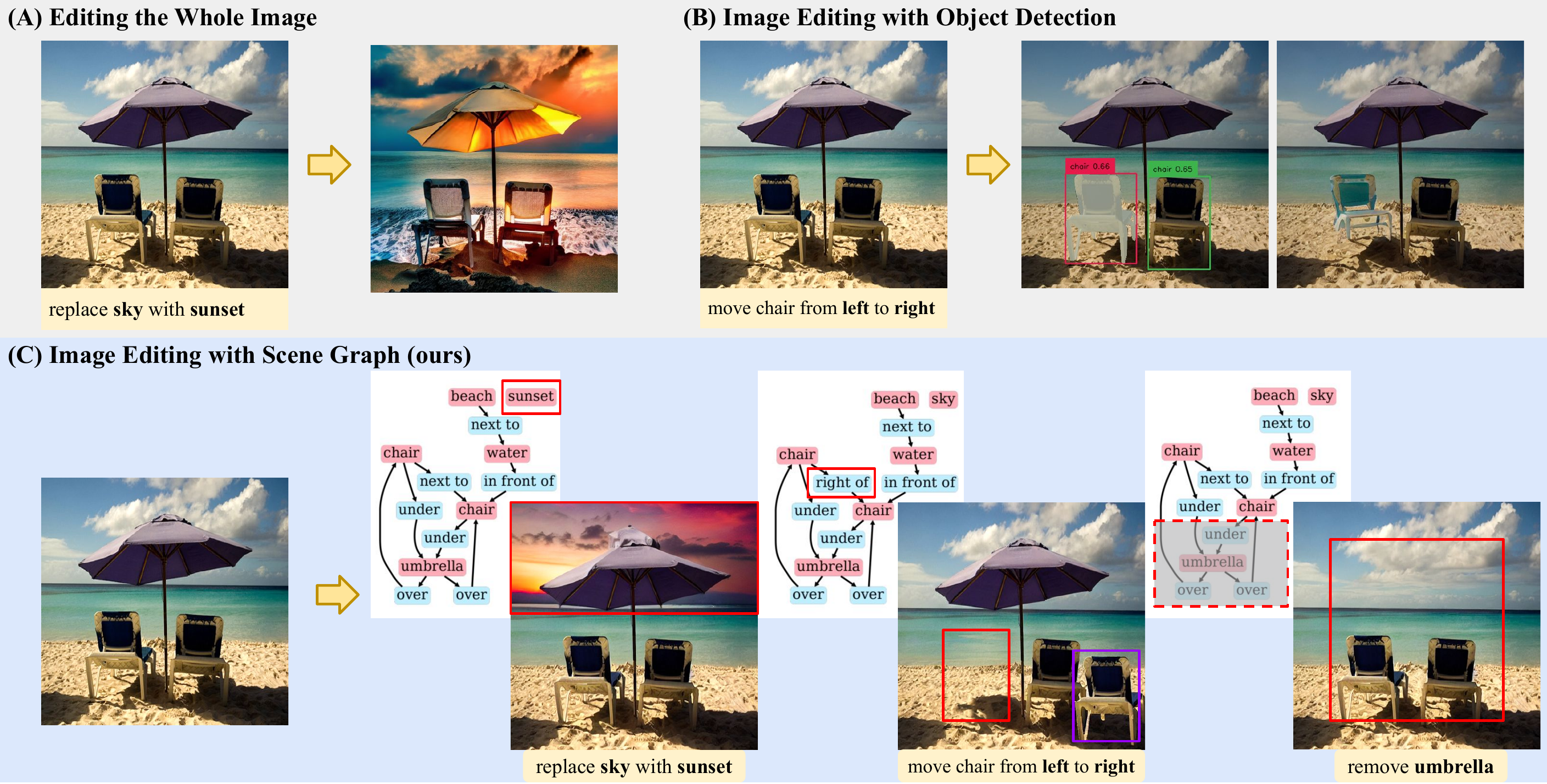}
    \vspace{-2mm}
    \caption{Prior work on text-to-image editing generally follow two approaches: (A), editing the entire image (\eg, Imagic~\cite{kawar2023imagic} or Dreambooth~\cite{ruiz2022dreambooth}), or (B), localizing the Region of Interest (RoI) through user-provided bounding boxes or object detection (\eg, Grounded-SAM~\cite{liu2023grounding,kirillov2023segment}+Stable Diffusion~\cite{rombach2022high}). In our work, shown in (C), we localize the RoI (outlined by the red bounding box) and predict the desired region (outlined by the purple bounding box) for the target object using a scene graph of an input image. This enables us to perform many editing operations in complex scenes, such as relationship change and object replacement, which were not supported by prior work.}
    \vspace{-4mm}
    \label{fig1: overview}
\end{figure}


Text-to-image editing aims to modify the specific content of an image based on language descriptions. Prior work (\eg,~\cite{rombach2022high,saharia2022photorealistic,nichol2021glide,bar2022text2live,kawar2023imagic}) either edits the input image globally, where the entire image is subject to modification~\cite{kawar2023imagic, li2020manigan}, or localizes the Region of Interest (RoI\footnote{We use RoI to indicate the region that is supposed to be edited in the image.}) according to either human-drawn~\cite{rombach2022high, nichol2021glide, avrahami2022blended} or detected bounding boxes~\cite{bar2022text2live}. These methods have two major drawbacks. First, accurately localizing and moving a target object in complex scenes can be challenging due to inherent ambiguities of text. For example, in Figure \ref{fig1: overview}(A), methods like Imagic~\cite{kawar2023imagic} edits the entire image instead of the target object ``sky.'' In Figure \ref{fig1: overview}(B), Liu \etal~\cite{liu2023grounding} randomly selects a chair since they have the same semantic meaning. The second major drawback we observe is that most existing image inpainting methods are trained using free-form masks~\cite{nazeri2019edgeconnect,yu2019free,rombach2022high}. As our experiments will show, these methods may fail to add or replace the target object according to prompts, especially when the RoI is small. These methods fill in the masked regions by guessing at the background, instead of inserting the desired object according to the text prompt. 

To address the aforementioned issues, we propose the Scene Graph Comprehension Network (SGC-Net) for complex scene image editing. In Figure \ref{fig1: overview}(C), where the desired edit is ``move the chair from left to right,'' the model must first localize the target object (the chair on the left side) given the text description. The model then must recognize the desired action of placing the chair to the right-hand side of the second chair.  SGC-Net accomplishes this by converting the desired edits into changes in a scene graph representation of the image.  Then, given the new scene graph, SGC-Net produces the edited image.  This approach aids SGC-Net in more accurately identifying the target object despite textual ambiguities than prior work (\eg,~\cite{liu2023grounding,kirillov2023segment,nichol2021glide,rombach2022high}), as well as helps ensure that the edited object has the correct relationship to other objects in the scene.

\begin{figure}[t]
    \centering
    \includegraphics[width=.75\linewidth]{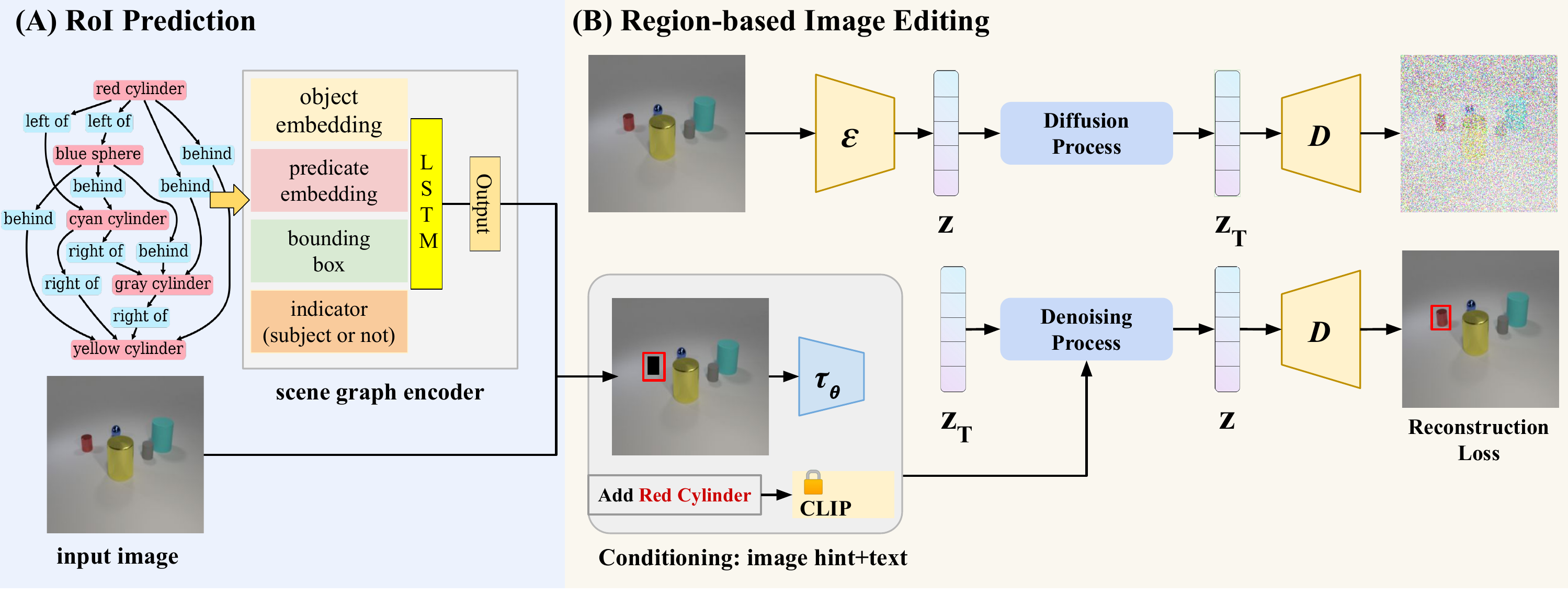}
    \vspace{-2mm}
     \caption{\textbf{SGC-Net overview.} Our approach consists of two sequential stages: (A) RoI prediction, which localizes the RoI based on a modified scene graph. Specifically, a scene graph encoder takes the modified scene graph as input and outputs a bounding box. (B) Region-based image editing: During inference, we take the image masked by predicted bounding boxes as input and outputs the modified image. During training, we randomly mask out objects to simulate the output of our RoI prediction module, and train our model to reconstruct the original image. Thus, our model does not require image pairs before and after editing. }
     \vspace{-4mm}
    \label{fig2: approach}
\end{figure}

More formally, given an image to be edited,  we extract editing triplets from the given text prompt, \eg, $<$target chair -- right of -- reference chair$>$,  that we use to make changes to the scene graph representing the input image. Rather than relying solely on text descriptions that require a model to implicitly infer any changes to the spatial information in an image, our model explicitly predicts the desired location of target object(s) based on their relationship to other objects encoded in the scene graph. For example, for the triplet $<$target chair -- right of -- reference chair$>$,  the first stage of our model uses an RoI prediction module to identify that the desired region for the target chair should be on the right side of the reference chair (illustrated in Figure \ref{fig2: approach}(A)). Then, in the second stage SGC-Net edits the image conditioned on the text prompt and RoI predictions from the first stage (shown in Figure \ref{fig2: approach}(B)).

In summary, the contributions of this paper are:
\begin{itemize}[nosep,leftmargin=*]
\item We propose a Scene Graph Comprehension Network (SGC-Net) that performs text-to-image editing using scene graphs, reducing manual effort and alleviating the issues caused by the ambiguity of text. Compared to methods that require pre-defined masks or bounding boxes as input~\cite{nichol2021glide,rombach2022high}, our model uses the semantic meanings of objects and their relationships in an image to accurately predict desired RoI for the target object.
\item We propose a region-based image editing approach built on Stable Diffusion~\cite{rombach2022high} that can achieve different image editing tasks without paired training images. As we will show, our approach can even remove or add objects with small bounding boxes.
\item Our experiments demonstrate SGC-Net outperforms the state-of-the-art. \Eg, SGC-Net provides an 8 point boost in SSIM (RoI) on CLEVR~\cite{johnson2017clevr} and human users preferred SGC-Net image edits by 9-33\% on Visual Genome~\cite{krishna2017visual} over the baselines~\cite{dhamo2020semantic,rombach2022high,nichol2021glide}.
\item We perform experiments on in-the-wild\footnote{Following~\cite{zhang2022sine}, we refer to images sourced from the real world as "in-the-wild", as opposed to image datasets collected by humans.} image editing, validating the generalization and flexibility of our method.
\end{itemize}

\section{Related Work}
\label{sec:related_work}
\noindent \textbf{Text-to-Image Synthesis \& Editing}. Early studies in text-to-image synthesis and editing often relied on conditional GANs~\cite{xu2018attngan,zhang2018stackgan++,li2020manigan,liang2018generative,fan2023target}, but often were limited to generating low-resolution images due to high GPU memory requirements and scalability limitations of these methods. More recent methods (\eg,~\cite{ramesh2021zero,ramesh2022hierarchical,rombach2022high,saharia2022photorealistic,zhang2023adding}) have focused on training conditional diffusion models for text-to-image generation on large scale datasets (\eg, LAION-400M~\cite{schuhmann2021laion}). Building on these text-to-image generation frameworks, different editing approaches~\cite{bar2022text2live,nichol2021glide,ruiz2022dreambooth,kawar2023imagic,gal2022image} have been proposed to support image editing tasks such as attribute manipulation and image inpainting. These methods can be roughly divided into two major categories. First, methods that perform global image edits based on text prompts, such as Dreambooth~\cite{ruiz2022dreambooth}, Imagic~\cite{kawar2023imagic}, and Text Inversion~\cite{gal2022image}. They treat the entire image as the target object and manipulate the input image accordingly. The second category of methods~\cite{nichol2021glide,bar2022text2live,zhang2022semantic,rombach2022high} use region-based editing that only modifies the Region of Interest (RoI). These methods obtain the target region using either user-provided masks~\cite{nichol2021glide,rombach2022high} or those from an object detector~\cite{bar2022text2live,liu2023grounding}. While these methods work well for images with few objects and simple semantic relationships, as we have discussed, accurately locating and manipulating target objects in complex scenes (\eg, those with multiple object that have similar semantic meanings) can be challenging. To address this, our SGC-Net utilizes a scene graph encoder to accurately localize the RoI by taking advantage of object relationships and predict the desired region for the target object without requiring user-provided masks. 
\smallskip

\noindent \textbf{Scene Graphs and Visual Relationship Detection.} Scene graphs~\cite{johnson2015image} describe the objects, attributes of objects, and relationships between objects in an image, and methods that generate them can be divided into two categories: Convolutional Neural Network (CNN)-based methods~\cite{li2017scene, li2018factorizable, yang2018graph, qi2019attentive} and Recurrent Neural Networks (RNN)-based methods~\cite{herzig2018mapping, xu2017scene, zellers2018neural}. At a high level, scene graph construction combines object/entity detection~\cite{ren2015faster,plummer2020revisiting} and detecting their visual relationships~\cite{dai2017detecting, lu2016visual, plummer2017phrase}. Our paper also is related to tasks where they aim to perform entity localization according to visual relationships (\eg, \cite{krishna2018referring,plummer2017phrase}).  
However, unlike these tasks, in our setting the target object is not always visible in the image. Therefore, we propose an RNN-based method to automatically predict a plausible position for the target object according to the existing information from the image and $<$\texttt{subject-predicate-object}$>$ triplets.

\section{SGC-Net: Scene Graph Comprehension Network for image editing}
\label{sec:SGN-Net}
Given an input image $x$, we define the entities in $x$ as $O = \{o_1,...,o_n\}$, the corresponding bounding boxes as $B = \{b_1,...,b_n\}$, and the relationships between entities as $R=\{r_1,...,r_m\}$. Our task is to perform semantic manipulation on $x$ based on a scene graph $G$ that consists of $\{O, B, R\}$, where the scene graph can be directly modified by users or modified according to the user-provided text prompts. 
To manipulate the image $x$ according to the scene graph $G$, our SGC-Net uses a two step process.  First, Section \ref{sec:SGC-Net_RoI_prediction} describes our Region-of-Interest (RoI) prediction module that identifies the target manipulation predicting the desired regions for the target object with scene graph information $G$. Second, Section \ref{sec:SGC-Net_region_based_editing} presents our region-based image editing model, which aims to understand and implement various image editing operations. Finally, Section \ref{sec:SGC-Net_scene_graph_modification} describes the editing operations and their corresponding modifications on scene graph.  Figure \ref{fig2: approach} provides an overview of our approach.

\subsection{Region of Interest (RoI) Prediction}
\label{sec:SGC-Net_RoI_prediction}
As discussed in the Introduction, correctly localizing and predicting the desired regions for the target object is critical for complex scene image editing. To address the text ambiguity issues in methods based on object detection~\cite{liu2023grounding,bar2022text2live}, we propose a RoI prediction module based on scene graph to facilitate various image editing tasks.  This module is especially useful for tasks like semantic relationship change and object addition since they explicitly identify the location to place the target object. This module enables SGC-Net to accurately localize the target object and predict where to place it, instead of randomly selecting one object that satisfies the semantic meaning as the target object. Since the target object may have relationships with multiple objects in a scene, we purpose an LSTM-based model that encodes all the triplets of the modified scene graph.





Given a modified scene graph $\widetilde{G}$ and the target object $o$ with triplets \textbf{y} = $\{y_{1}, ... ,y_{T}\}$. For $t\in\{1,...,T\}$, $y_{t}$ is a triplet in $<s-p-o>$ format, where $s,o\in O$ and $p\in R$. For each triplet $y_t$, we devise two embedding layers that obtain the object embedding $V_s$, $V_o$ and predicate embedding $V_p$ separately. We also introduce a binary indicator $I$ to indicate whether the target object is subject or object in $y_t$. Suppose the reference object corresponds to ``subject'' and the target object corresponds to ``object,'' we also consider the position of reference object $b_s$ as part of the input to our model. We concatenate these features and encode them with an LSTM. Specifically, it can be expressed as: 
\begin{equation}
    m_t = concat\{V_s, V_o, V_p, b_s, I\}; \hspace{4mm} h_t = LSTM(m_t,h_{t-1}) 
\end{equation}
\noindent where $h_t$ is the hidden state of LSTM at triplet $y_t$. We apply an Multilayer Perceptron (MLP) on $h_T$ to predict the four coordinates of RoI. Since the image size is normalized, we crop the final output to range(0$\sim$1) and train the model using Mean Squared Error. Ss shown in Figure \ref{fig2: approach}, once we obtain the predicted RoI, we generate a masked image $x_{hint}$, where the RoI is masked out and given as input to the image editing module described in the next section.

\subsection{Region-based Image Editing}
\label{sec:SGC-Net_region_based_editing}

After obtaining the predicted bounding boxes from our RoI prediction module described in Section~\ref{sec:SGC-Net_RoI_prediction}, a straightforward approach to image editing is to apply an image inpainting method, such as GLIDE~\cite{nichol2021glide} or Stable Diffusion~\cite{rombach2022high}, directly to the masked image $x_{hint}$. However, as our experiments will show, these approach struggles to accurately add or remove objects in complex scenes, particularly when the bounding boxes are small, or distinguish between operations like addition and removal. 
For example, when asked to remove an object these methods tend to add the object to the modified region instead. This is due, in part, to the fact that many models (\eg,~\cite{rombach2022high,saharia2022photorealistic}) were pretrained on datasets like LAION-400M~\cite{schuhmann2021laion} that contain generic image captions that likely do not contain editing-specific language like ``\textit{remove} a car'' or ``\textit{add} a car.'' Thus, during training we generate a text prompt using a template for each editing operation we support, and then train our region-based image editing module conditioned on our generated prompts. Additionally, rather than directly inpaint $x_{hint}$, we provide it as an additional control to our diffusion process. This enables SGC-Net to decide what parts are important to keep and what may be safely ignored.


More formally, our region-based image editing approach uses Stable Diffusion~\cite{rombach2022high} to reconstruct the original image $x$ from the masked image $x_{hint}$ according to user's input. During training $x_{hint}$ can be obtained by randomly removing objects from $x$, thereby eliminating the requirement for paired image data before and after editing.  
Specifically, given the image data $x_0$ and a randomly generated noisy image $x_T$, Stable Diffusion~\cite{rombach2022high} can be considered as a series of equally-weighted denoising autoencoders $\epsilon_{\theta}(x_t,t)$. Here $T$ represents the length of the Markov Chain in the diffusion process, and $t$ denotes the time step, ranging from $1$ to $T$. For unconditional generation, the denoising autoencoders are trained to predict the corresponding noise $\epsilon$ with an objective function:
\begin{align}
    L_{LDM} := \mathbb{E}_{\mathcal{E}(I),\epsilon\sim \mathcal{N}(0,1),t}\left[ || \epsilon - \epsilon_\theta(z_t,t) ||^2_2 \right],
\label{eq: ldm}
\end{align}
\noindent where $\mathcal{E}$ is the pretrained encoder of VQGAN~\cite{esser2021taming} to encode image $x_t$ to latent features $z_t$, and vice versa. For conditional generation, the denoising autoencoders $\epsilon$ take $\tau_{\theta}(y)$ as additional input and the Conditional Latent Diffusion Model (CLDM) is optimized by
\begin{align}
    L_{CLDM} := \mathbb{E}_{\mathcal{E}(I), y, \epsilon\sim \mathcal{N}(0,1),t}\left[ || \epsilon - \epsilon_\theta(z_t,t, \tau_{\theta}(y)) ||^2_2 \right],
\label{eq: conditional_ldm}
\end{align}
\noindent where $\tau_\theta$ and $\epsilon_\theta$ are optimized jointly. As discussed earlier, during training our text prompts are created from a template that include information about the type of editing operation being performed.
In addition, to perform editing operations around the bounding box (\eg, add shadows to the red cylinder in Figure \ref{fig2: approach}), we use $x_{hint}$, which is encoded by ControlNet~\cite{zhang2023adding}, as an image hint in our conditions, rather than just as input to the denoising autoencoders $\epsilon_{\theta}(x_t,t)$ like GLIDE~\cite{nichol2021glide} or Stable Diffusion~\cite{rombach2022high}. This allows our model to perform various editing tasks both within and around the RoI, resulting in more visually natural outputs. 




\subsection{Image Editing Operations}
\label{sec:SGC-Net_scene_graph_modification}
Following~\citet{dhamo2020semantic}, we perform four image manipulation tasks: object addition, object replacement, relationship change, and object removal\footnote{Following \citet{dhamo2020semantic}, we perform all four operations on CLEVR~\cite{johnson2017clevr}, and three operations on Visual Genome~\cite{krishna2017visual}: object replacement, relationship change, and object removal.}. These manipulations are reflected on the nodes and edges of scene graphs. Take the red cylinder in Figure \ref{fig2: approach} as an example, (1) Object Addition: adding a new object (node) and its corresponding relationships (edges) with other objects on the scene graph; (2) Object Replacement: replacing the node which represents $<$\texttt{red cylinder}$>$ to another object; (3) Relationship Change: changing the spatial relationships of red cylinder (edge) from $<$\texttt{red cylinder-left of-blue sphere}$>$ to $<$\texttt{red cylinder-right of-blue sphere}$>$. Other relationships of the red cylinder can be changed similarly; (4) Object Removal: deleting the node of $<$\texttt{red cylinder}$>$ and its corresponding edges.
\smallskip

\section{Experiments}
\label{sec:experiments}


\noindent \textbf{Datasets.} We evaluate the performance of our model on CLEVR~\cite{johnson2017clevr} and Visual Genome~\cite{krishna2017visual}. CLEVR is a synthetic dataset that contains ground truth pairs for image editing. We use the test set provided by~\citet{dhamo2020semantic} for a fair comparison to prior work. However, Visual Genome, lacks image pairs before and after editing. Thus, we evaluate performance using human judgements to estimate the correctness of manipulation. 
\smallskip

\noindent \textbf{Metrics.} Following~\citet{dhamo2020semantic}, we report mean absolute error (MAE) and structural similarity index measure (SSIM) of RoI as the evalution metrics. The RoI is defined as the area of image that should be modified according to user's input, which can directly reflect the accuracy of the manipulation. 
\smallskip

\noindent \textbf{Baselines.}
We choose SIMSG~\cite{dhamo2020semantic}, GLIDE~\cite{nichol2021glide}, and Stable Diffusion (SDM) v2~\cite{rombach2022high} as our baselines for image editing. Since GLIDE and SDM require user-provided masks as input, we use the bounding boxes predicted by our RoI prediction module as input to GLIDE and SDM on CLEVR. For Visual Genome, we use Grounded-SAM~\cite{liu2023grounding,kirillov2023segment} to automatically detect the segmentation masks for GLIDE and SDM\footnote{We refrain from using Grounded-SAM on CLEVR since its images always consist of multiple semantically similar objects, making it difficult for Grounded-SAM to accurately localize the target object.}.
\smallskip

\noindent \textbf{Implementation Details.}
SGC-Net has two stages: the RoI prediction and region-based image editing. We train the two stages separately. To train the RoI prediction module, we extract modified scene graphs as input and the bounding boxes of target objects as output. We set the maximum number of the modified triplets of each image to 5. The mean absolute error (MAE) on the validation set of CLEVR is 12.05$\pm$0.87 (computed over 5 runs), with images of size 256$\times$256. The MAE on the Visual Genome dataset is 77.84$\pm$1.25 (computed over 5 runs), with images of size 512$\times$512. For the region-based editing model, we simulate the output $x_{hint}$ of our RoI prediction model by randomly masking out objects. We use $x_{hint}$ as input and the original image $x$ as output. Finally, we adopt a compositional scene representation mechanism following~\citet{zhang2022semantic} to preserve the original content of unmodified regions. To eliminate obvious artificial effects in the boundary area, we apply Possion Blending~\cite{perez2003poisson} to combine the input image and the modified regions.


\subsection{Results on CLEVR}
\label{sec:experiments_CLEVR}

\begin{figure}[t]
    \centering
    \includegraphics[width=0.7\linewidth]{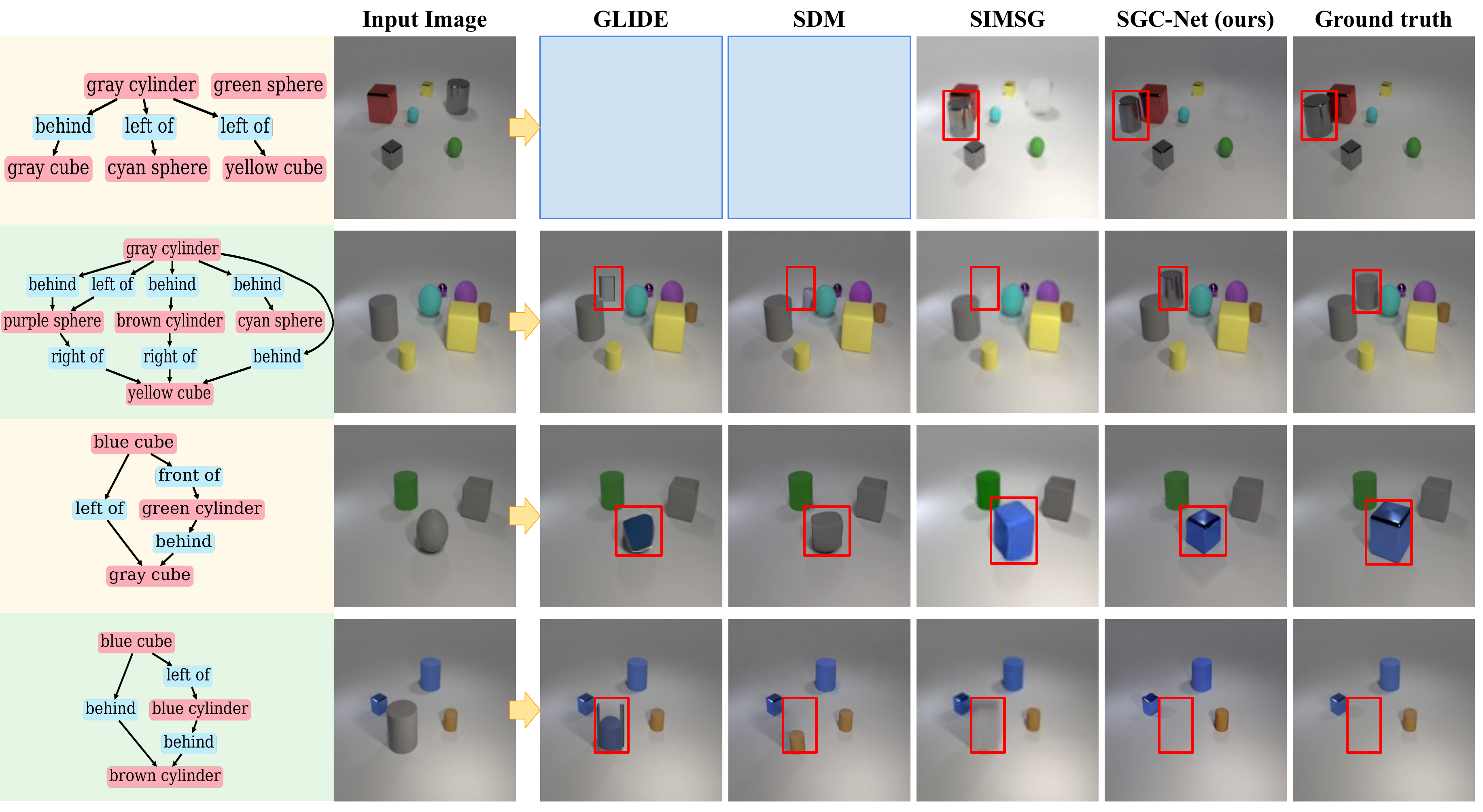}
    \vspace{-2mm}
    \caption{Qualitative examples for editing 256$\times$256 images on CLEVR. Tasks from top to bottom: semantic relationship change, object addition, object replacement, and object removal. Blank image means the corresponding approach is not capable of this task. We outline the RoI by red bounding boxes. We observe that SGC-Net can accurately predict the RoI and edit objects in complex scenes. See Section \ref{sec:experiments_CLEVR} for discussion.} 
    \vspace{-2mm}
    \label{fig:clevr_qualitative}
\end{figure}

We present quantitative and qualitative results on 256$\times$256 images\footnote{We apply super-resolution method USRNet~\cite{zhang2020deep} on the output of SIMSG (128$\times$128 on CLEVR and 64$\times$64 on Visual Genome) to avoid out-of-memory issues on a 48GB GPU.} in Table \ref{table:clevr_quantitative} and Figure \ref{fig:clevr_qualitative}, respectively. We draw three major conclusions. First, text-to-image inpainting methods~\cite{nichol2021glide,rombach2022high} pretrained on massive datasets do not perform well on our image editing tasks. As discussed in Section \ref{sec:SGC-Net_region_based_editing}, they tend to introduce unexpected objects to repair the missing regions. This shown by the significant performance gains on object removal reported in Table \ref{table:clevr_quantitative} and also illustrated in the removal example in Figure \ref{fig:clevr_qualitative}, where GLIDE uses a distorted blue cube to fill the missing regions and SDM draws the yellow cylinder from the right-hand side to inpaint the masked region. Our second conclusion is that text-driven localization methods struggle to accurately localize the RoI and desired regions. \Eg, there are two grey cylinders and two purple spheres for image addition example in Figure \ref{fig:clevr_qualitative}. Even a perfect object detection method only has a 25\% chance of correctly localize the target objects, let alone understand the semantic relationships between them. Finally, our third observation notes that although SIMSG achieves comparable SSIM scores with our method on object replacement and addition, the MAE loss of SIMSG is much higher than ours, which means some attributes like brightness may be modified by SIMSG (\eg, as shown in Figure \ref{fig:clevr_qualitative}).

\begin{table}[t]
\centering
\setlength{\tabcolsep}{2.pt}
\begin{tabular}{l cc cc cc cc}
\toprule
\multirow{2}*{\textbf{Method}} &
\multicolumn{2}{c}{\textbf{Object}} &
\multicolumn{2}{c}{\textbf{Object}} & 
\multicolumn{2}{c}{\textbf{Object}} & 
\multicolumn{2}{c}{\textbf{Relationship}} \\
& \multicolumn{2}{c}{\textbf{Removal}} & \multicolumn{2}{c}{\textbf{Replacement}} & \multicolumn{2}{c}{\textbf{Addition}} & \multicolumn{2}{c}{\textbf{Change}}\\
& MAE $\downarrow$ & SSIM $\uparrow$ 
& MAE $\downarrow$ & SSIM $\uparrow$ 
& MAE $\downarrow$ & SSIM $\uparrow$ 
& MAE $\downarrow$ & SSIM $\uparrow$ 
\\
\midrule
GLIDE~\cite{nichol2021glide}+RoI & 39.51 & 62.91 & 34.59 & 55.57 & 40.25 & 59.04 & - & - \\
SDM v2~\cite{rombach2022high}+RoI & 31.48 & 72.65 & 36.49 & 54.48 & 44.39 & 57.61 & - & - \\
SIMSG~\cite{dhamo2020semantic} & 30.38 & 85.89 & 33.76 & 67.41 & 44.58 & \textbf{65.80} & 33.31 & 85.95 \\
SGC-Net (ours) & \textbf{9.90} & \textbf{93.33} & \textbf{27.64} & \textbf{68.05} & \textbf{34.18} & 64.78 & \textbf{11.40} & \textbf{92.32} \\
\bottomrule
\end{tabular}
\caption{Quantitative results for editing 256$\times$256 images on CLEVR. Omitted results indicate a method that cannot support a task. Since GLIDE and SDM requires user-provided inpainting mask, we use the predicted bounding boxes by our method as input, which is denoted by ``RoI''. SGC-Net notably outperforms baselines, especially for the object removal and relationship change tasks. See Section \ref{sec:experiments_CLEVR} for discussion.}
\vspace{-4mm}
\label{table:clevr_quantitative}
\end{table}

\subsection{Results on Visual Genome}
\label{sec:experiments_VG}

As with experiments on CLEVR, we apply GLIDE and SDM on object removal and object replacement tasks, and SIMSG on all editing tasks. For each image editing task, we randomly selected 30 images generated by each baseline. This resulted in 120 images for object removal and replacement, and 60 images for semantic relationship change. Each image was annotated three times by AMT workers where they were asked to judge whether the image is correctly manipulated according to the input guidance. In Table \ref{table:vg_quantitative}, we report that SGC-Net significantly outperforms baselines, especially for object removal and relationship change tasks, which is consistent with our observation on CLEVR.

\begin{table}[t]
\centering
\begin{tabular}{l c c c}
\toprule
& Object & Object & Relationship \\
& Removal  & Replacement  & Change \\
\midrule
GLIDE~\cite{nichol2021glide} & 23.3\% & 28.8\% & - \\
Stable Diffusion v2~\citep{rombach2022high} & 21.1\% & 32.2\% & - \\
SIMSG~\citep{dhamo2020semantic} & 22.2\% & 24.4\% & 15.6\%\\
SGC-Net(ours) &  \textbf{50.0\%} & \textbf{40.0\%} & \textbf{48.9\%} \\
\bottomrule
\end{tabular}
\caption{User judgments on the correctness of an image manipulation on Visual Genome. Empty values indicate the approach is not capable of this task. SGC-Net outperforms baselines in image manipulation accuracy, especially for object removal and relationship change, aligning with our observation on CLEVR results. See Section \ref{sec:experiments_VG} for discussion.}
\label{table:vg_quantitative}
\vspace{-2mm}
\end{table}

\begin{figure}[t]
    \centering
    \includegraphics[width=0.75\linewidth]{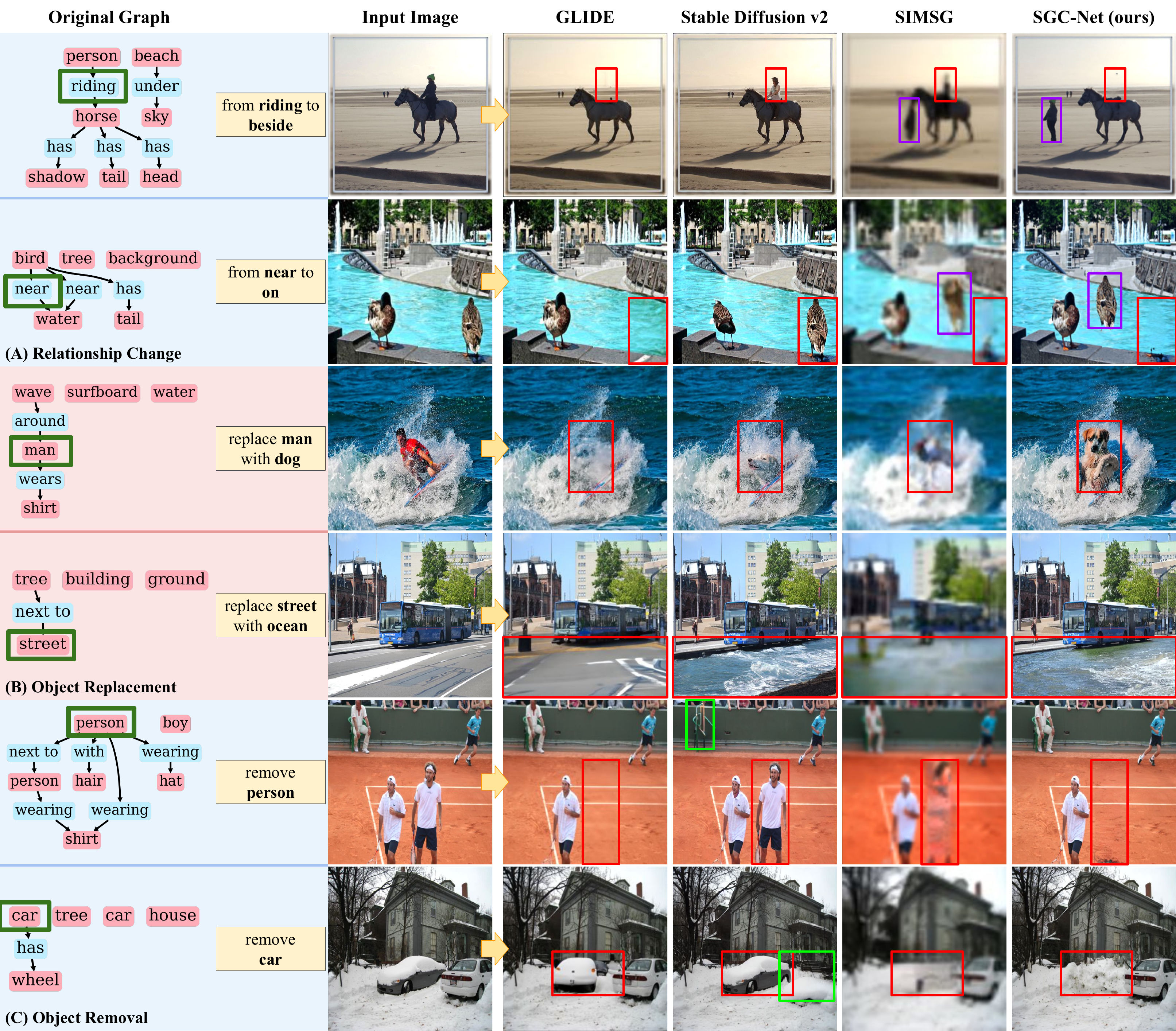}
    \vspace{-2mm}
     \caption{Qualitative examples for editing 512$\times$512 images on Visual Genome. The original region of the target object, desired region for the target object are outlined by red bounding box and purple bounding box respectively.  We simplify the scene graphs for better visualiation. SGC-Net not only predicts desired regions for the target objects but also edits these regions based on user modifications better than prior work. See Section \ref{sec:experiments_VG} for discussion.} 
     \vspace{-2mm}
    \label{fig:vg_qualitative}
\end{figure}

We present qualitative examples in Figure \ref{fig:vg_qualitative}. Figure \ref{fig:vg_qualitative}(A) shows that our RoI prediction module outputs reasonable desired regions for the target objects when changing its relationship to other objects in the scene, such as ``person beside horse'' and ``bird on water''. In contrast, methods such as GLIDE or SDM do not understand the semantic relationship change between objects and cannot edit the image accordingly. Figure \ref{fig:vg_qualitative}(B) presents object replacement examples, where SGC-Net outperforms both the scene-graph driven approach SIMSG and the text-driven approaches GLIDE and BDM. For instance, GLIDE often guesses at the background without properly inserting the object into the masked area. Figure \ref{fig:vg_qualitative}(C) shows object removal examples. To demonstrate the effectiveness of our RoI prediction module, we apply the red bounding boxes predicted by our model as input to GLIDE and the green bounding boxes predicted by Grounded-SAM~\cite{liu2023grounding, kirillov2023segment} as input to SDM. It can be observed that without the input masks predicted by our model, SDM failed to accurately localize the target object due to the presence of multiple "cars" and "people" in the images. In contrast, GLIDE correctly localized the target object according to the bounding boxes provided by our method. However, as discussed in Section \ref{sec:SGC-Net_region_based_editing}, methods like GLIDE may introduce unexpected objects to the image, as seen in the second object removal example where GLIDE introduces a new car covered by snow to inpaint the target region. Our region-based editing model, on the other hand, does not suffer from this issue due to its training strategy.

\subsection{In-the-wild Text-to-image Editing}
\label{sec:experiments_in_the_wild}
\begin{figure}[t]
    \centering
    \includegraphics[width=.75\linewidth]{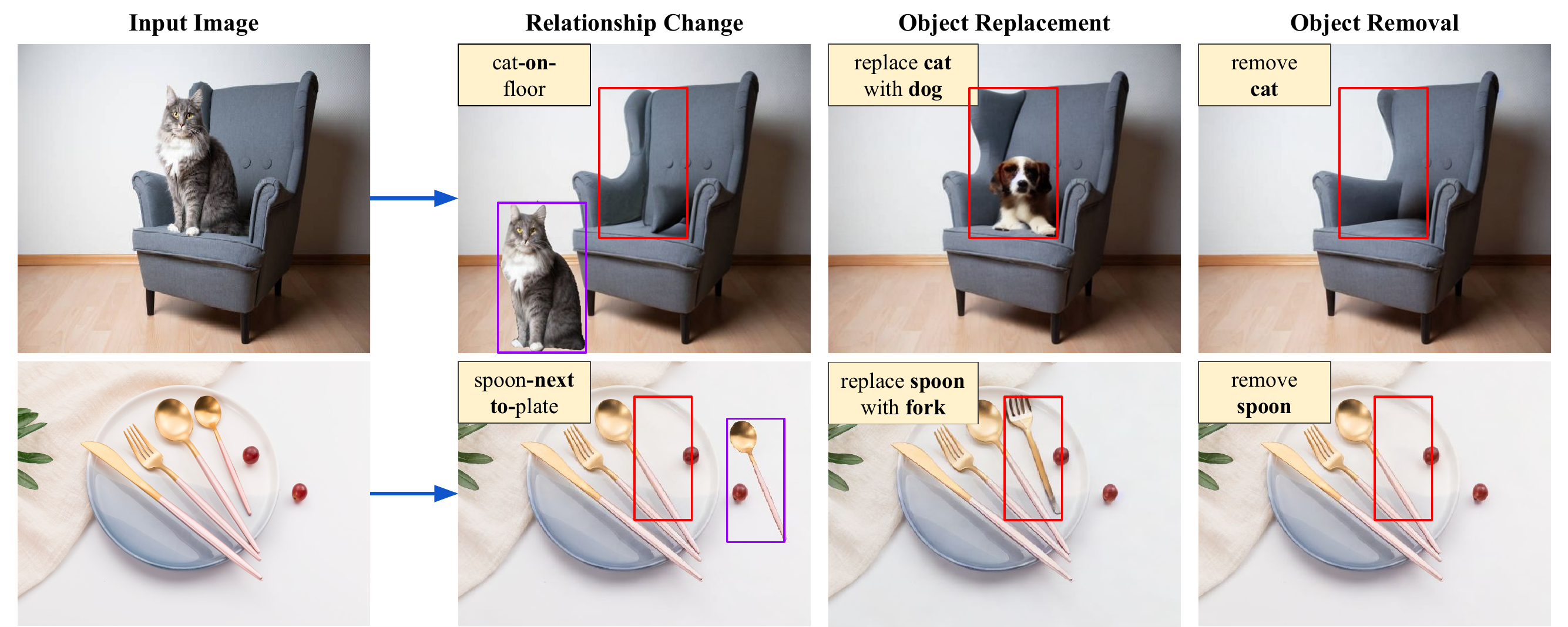}
    \vspace{-2mm}
    \caption{In-the-wild text-to-image editing results. SGC-Net can be adapted to text prompts with localization methods (\eg, Grounded-SAM~\cite{liu2023grounding, kirillov2023segment}). See Section \ref{sec:experiments_in_the_wild} for discussion.}
    \vspace{-2mm}
    \label{fig:in-the-wild}
\end{figure}

Our experiments have demonstrated that incorporating scene graph information can significantly improve the accuracy of image editing compared to using text alone. While modifying the scene graph incurs additional overhead compared to editing text only, we demonstrate that our model can be easily adapted to text commands in this section. We show diverse generated results in Figure \ref{fig:in-the-wild}, where our model accurately edits the target object according to the $<$\texttt{subject-predicate-object}$>$ triplets that are converted from text commands. For instance, the cat has been moved from the sofa to the floor, and the spoon has been moved from inside the plate to beside the plate.

\subsection{Limitations and Future Work}
\label{sec:limitations}
\begin{figure}[t]
    \centering
    \includegraphics[width=0.75\linewidth]{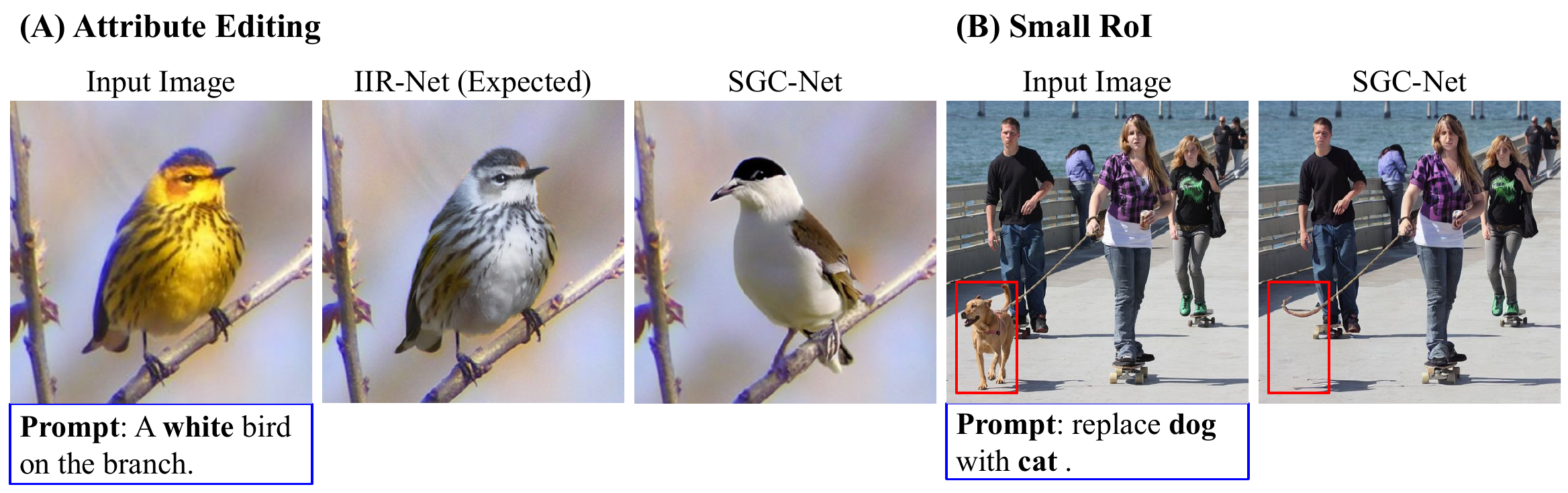}
    \vspace{-2mm}
    \caption{Limitations of SGC-Net. (A) Attribute editing: SGC-Net lacks the ability to modify object attributes. (B) Inserting objects in small RoI: While SGC-Net can effectively alleviate the background guessing issue, we still observe some failure cases on Visual Genome when the RoI is small. See Section \ref{sec:limitations} for discussion.}
    \vspace{-2mm}
    \label{fig:limitations}
\end{figure}

We find SGC-Net can edit objects in complex scenes, but struggles to edit object attributes. \Eg, as shown in Figure \ref{fig:limitations}(A), we could replace yellow bird with a white bird, but cannot produce the same bird with a different attribute. Thus, exploring ways to integrate attribute editing methods(\eg, Imagic~\cite{kawar2023imagic} or IIR-Net~\cite{zhang2023text}), to generate objects with different attributes is a good direction for future work. 
Additionally, while our method effectively alleviates the issue of inserting objects into small regions, we still observe some failure cases where it may struggle to insert the objects, particularly in some cases on Visual Genome shown in Figure \ref{fig:limitations}(B). Based on this finding, collecting more training images containing small objects may further improve the performance on object editing for small RoIs.

\section{Conclusion}
In this paper, we propose SGC-Net, a semantic image editing method that uses scene graph information to achieve complex scene image editing. SGC-Net consists of two stages: an RNN-based RoI prediction module that identifies regions to edit the target objects, and a region-based image editing module that supports many editing tasks such as object replacement, removal, and relationship change. We demonstrate that SGC-Net outperforms the state-of-the-art in both qualitative and quantitative evaluations on CLEVR and Visual Genome. For example, SGC-Net achieves around an 8-point gain in SSIM on CLEVR and a 9-33\% boost in user evaluation accuracy on Visual Genome. Furthermore, experiments on in-the-wild image editing demonstrate SGC-Net's ability to generalize to practical settings.

\noindent\textbf{Acknowledgements.} This material is based upon work supported, in part, by DARPA under agreement number HR00112020054. Any opinions, findings, and conclusions or recommendations expressed in this material are those of the author(s) and do not necessarily reflect the views of the supporting agencies.

\bibliography{egbib}

\appendix
\section*{Appendices}

\section{Additional Experiment Results}
\label{sec:additional_results}

\subsection{Qualitative Results.} 
\begin{figure}[t]
    \centering
    \includegraphics[width=1.\linewidth]{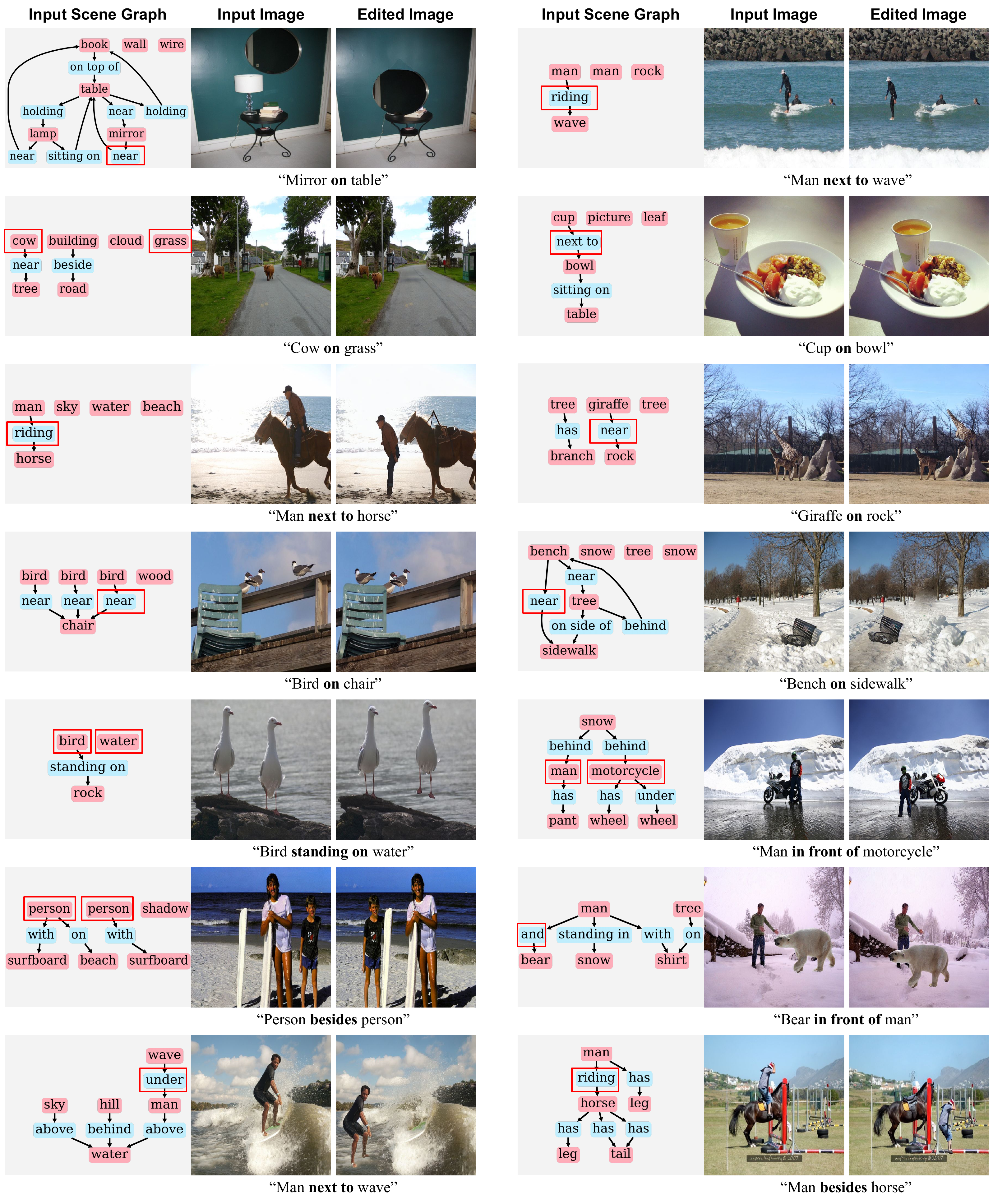}
    \caption{\textbf{Semantic Relationship Change.} Additional results of SGC-Net on the Visual Genome dataset. The modified nodes in scene graphs are outlined by red bounding boxes. We set the image resolution to 512$\times$512 and simplify the scene graphs for better visualization. See Appendix \ref{sec:additional_results} for discussion.} 
    \label{fig:supplementary_resposition}
\end{figure}

\begin{figure}[t]
    \centering
    \includegraphics[width=1.\linewidth]{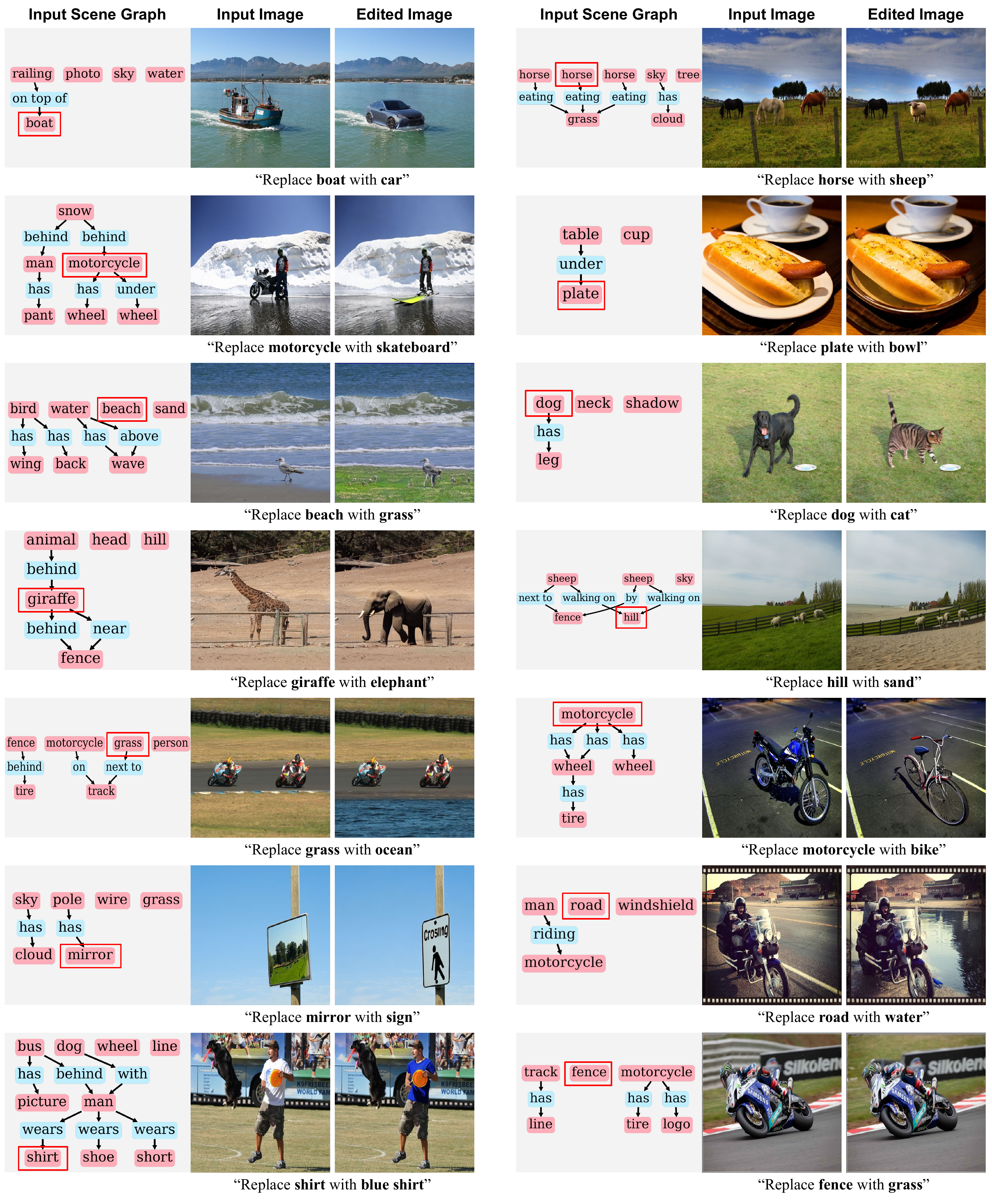}
    \caption{\textbf{Object Replacement.} Additional results of SGC-Net on the Visual Genome dataset. See Appendix \ref{sec:additional_results} for discussion.} 
    \label{fig:supplementary_replace}
\end{figure}

\begin{figure}[t]
    \centering
    \includegraphics[width=1.\linewidth]{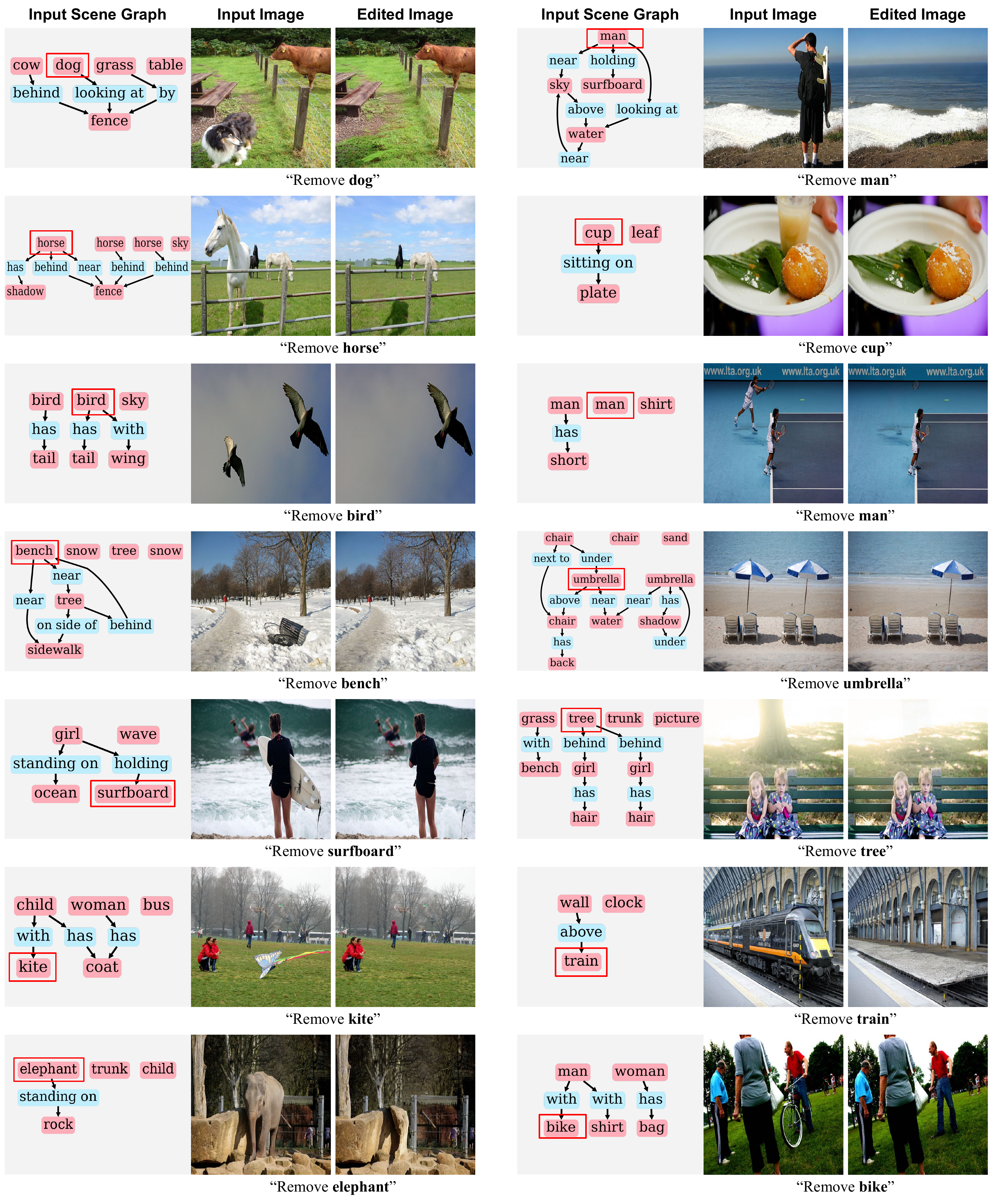}
    \caption{\textbf{Object Removal.} Additional results of SGC-Net on the Visual Genome dataset. See Appendix \ref{sec:additional_results} for discussion.} 
    \label{fig:supplementary_removal}
\end{figure}

We present additional qualitative results in Figure \ref{fig:supplementary_resposition}, \ref{fig:supplementary_replace}, and \ref{fig:supplementary_removal} to supplement the main paper. The results demonstrate the effectiveness of SGC-Net in performing various editing tasks based on modified scene graphs. For example, in Figure \ref{fig:supplementary_resposition}, we see that SGC-Net produces plausible regions for the target object when the semantic relationships have been changed, such as ``mirror -- on -- table'', ``man -- next to -- wave''. The observation is consistent with our conclusion in the main paper.
\smallskip

\subsection{Ablation Experiments on CLEVR.} Table \ref{table:ablation_study} ablates our two modules. We find a significant gain in scene graph comprehension compared to text-only RoI prediction (71.50$\rightarrow$79.48 on SSIM). In addition, our region-based editing module also boosts SDM (74.94$\rightarrow$79.48), validating the effectiveness of our proposed modules.

\begin{table}[ht]
    \centering
    \begin{tabular}{lcc}
    \hline
    Method & MAE(RoI)$\downarrow$ & SSIM(RoI)$\uparrow$ \\
    \hline
    SGC-Net(TEXT) & 27.28 & 71.50 \\
    SGC-Net(SDM) & 21.72 & 74.94 \\
    SGC-Net & \textbf{18.86} & \textbf{79.48} \\
    \hline
    \end{tabular}
    \caption{\textbf{Ablation study on CLEVR.} ``TEXT'' denotes text-only RoI prediction. ``SDM'' denotes Stable Diffusion [28].}
    \label{table:ablation_study}
\end{table}

\clearpage

\section{User study Template}
\label{sec:user_study_template}
\begin{figure}[h]
    \centering
    \includegraphics[width=.9\linewidth]{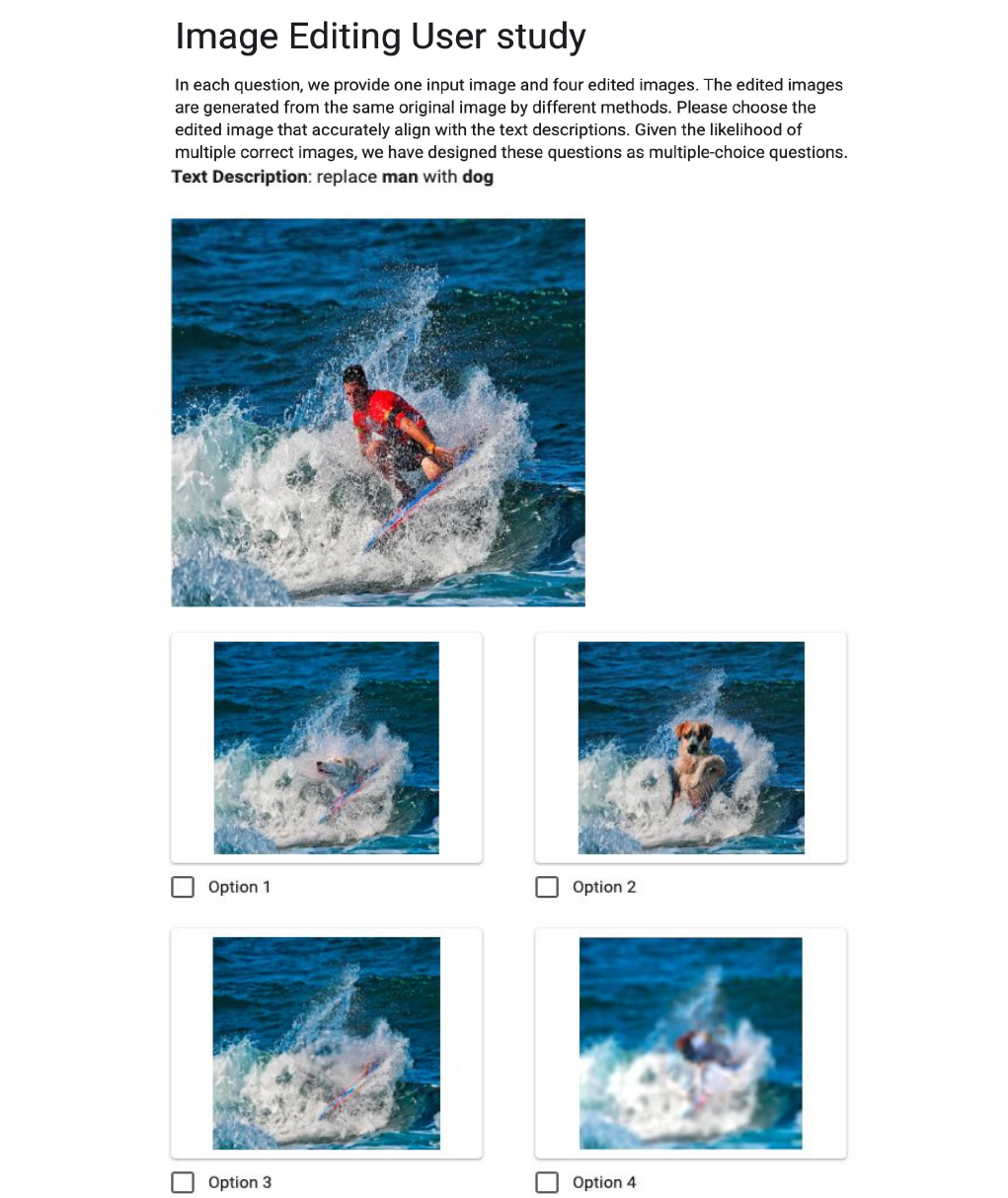}
     \caption{\textbf{User study screenshot.} A sample screenshot illustrating one of the questions presented to participants in our user study. See Appendix \ref{sec:user_study_template} for discussion.} 
    \label{fig:supplementary_user_study_template}
\end{figure}

In our user study, the annotators were shown an input image, a target text, and four edited images generated by different methods. The annotators were asked to choose which images accurately align with the target text. We provide a sample screenshoot in Figure \ref{fig:supplementary_user_study_template}.


\end{document}